\documentclass[11pt,a4paper]{article}

\usepackage[deanonymize]{tacl2021v1}

\usepackage{times}
\usepackage{latexsym}

\usepackage[T1]{fontenc}

\usepackage[utf8]{inputenc}

\usepackage{microtype}
\usepackage{lscape}

\usepackage{inconsolata}

\renewenvironment{quote}
               {\list{}{\rightmargin=0.1cm\leftmargin=0.3cm}%
                \item\relax}
               {\endlist}

\usepackage{graphicx} 
\usepackage{colortbl}
\usepackage{multirow}
\usepackage{hhline}
\usepackage{rotating}

\newcommand{\citeg}[1]{\citep[e.g.,][]{#1}}

\title{LLMs' Reading Comprehension Is Affected by Parametric Knowledge\\ and Struggles with Hypothetical Statements}

\author{Victoria Basmov\textsuperscript{\normalfont1,2} \, Yoav Goldberg\textsuperscript{\normalfont1,2} \,Reut Tsarfaty\textsuperscript{\normalfont 1}\\
\textsuperscript{1}Bar-Ilan University \, \textsuperscript{2}Allen Institute for Artificial Intelligence \\ 
{\tt\{\href{mailto:vikasaeta@gmail.com}{vikasaeta}, \href{mailto:yoav.goldberg@gmail.com}{yoav.goldberg},
\href{mailto:reut.tsarfaty@gmail.com}{reut.tsarfaty}\}
@gmail.com}}

\begin{document}
\maketitle
\begin{abstract}
The task of {\em reading comprehension} (RC), often implemented as {\em context-based question answering} (QA), provides a primary means to assess language models' natural language understanding (NLU) capabilities. Yet, when applied to  large language models (LLMs) with extensive built-in world knowledge, this method can be deceptive. If the context \emph{aligns} with the LLMs' internal knowledge, it is hard to discern whether the models' answers stem from context comprehension or  from  LLMs' internal information. Conversely, using data that \emph{conflicts} with the models' knowledge creates erroneous trends which distort the results. 
To address this issue, we suggest to use RC on \emph{imaginary data}, based on fictitious facts and entities. This task
is entirely independent of the models' world knowledge, enabling us to evaluate LLMs' linguistic abilities without the interference of parametric knowledge.
Testing ChatGPT, GPT-4, LLaMA 2 and Mixtral on such imaginary data, we uncover a class of linguistic phenomena posing a challenge to current LLMs,
involving
thinking in terms of {\em alternative, hypothetical} scenarios. While all the models handle simple affirmative and negative contexts with high accuracy, they are much more prone to error when dealing with {\em modal} and {\em conditional} contexts.
Crucially, these phenomena also trigger the LLMs' vulnerability to knowledge-conflicts
{again}.
In particular, while some models prove virtually unaffected by knowledge conflicts in affirmative and negative contexts, when faced with more semantically involved modal and conditional environments, they often fail to separate the text from their internal knowledge.

\end{abstract}

\section{Introduction}

A major mode of operation for large language models (LLMs) consists of following instructions that operate on a text provided in a user prompt. We call this operation mode {\em text-grounded prompting}, in contrast to {\em parametric prompting} where the LLM is expected to answer a user query based on its internal (parametric) knowledge, which has been  acquired during pre-training.
Text-grounded prompting is particularly important in situations where we seek outputs which are grounded in a specific text, such as in information extraction, claim verification or retrieval augmented generation (RAG), where the premise is that the provided text is more authoritative or more up-to-date than the LLM's internal parametric knowledge.

To perform well on the text-grounded scenario, the LLM must adhere to two requirements: (1) {\em understanding} the text in the prompt; and (2) being \emph{context-faithful}: answering exclusively based on information provided in the text, and not based on information in the parametric knowledge. 

There is a growing body of literature suggesting that LLMs are often {\em not} context-faithful, and that they answer while considering their parametric knowledge rather than relying exclusively on the text, in particular when the parametric knowledge conflicts with the text  \citep{DBLP:journals/corr/abs-2109-05052, neeman2022disentqa, li2022large, zhou2023contextfaithful,DBLP:journals/corr/abs-2305-12096,kasai2024realtime}. As we show in \S\ref{imaginary}, Figure \ref{fig:affirmatives}, there have been apparent advances, and GPT-4, LLaMa and Mixtral seemingly work well in relying exclusively on the context text, even in the presence of such knowledge conflicts.

In this work we are interested in property (1), the  ability to understand text, where we empirically measure ``text understanding'' through the task of reading comprehension: the ability to correctly answer questions based on the given text. Crucially, measuring understanding must take into account the interactions with parametric knowledge, as we want to ensure that the model's correct answer is based on the context text, and that the model is not ``cheating its way'' out of understanding the text by using its parametric knowledge. While previous work suggested the use of \emph{counterfactual instances}, where the context is changed to conflict with the LLMs parametric knowledge \citep{DBLP:journals/corr/abs-2109-05052,neeman2022disentqa,li2022large,zhou2023contextfaithful,Xie2023AdaptiveCO,DBLP:conf/emnlp/ChenZC22}, we argue that this is insufficient for assessing NLU  capabilities, as counterfactual instances result in knowledge conflicts between the text and the parametric knowledge, and the answer can be affected by these conflicts. Instead we propose to use \emph{imaginary instances}, that is, RC instances that use fictitious entities and facts, that are neither supported by nor contradict the parametric knowledge. These instances contain contexts and questions about made-up entities that do not refer to any  real-world entity, and do not have intersections with the LLMs' internal knowledge (\S\ref{imaginary}).
Having neutralized parametric knowledge effects, we are ready to proceed to NLU assessment. 

Most previous works on NLU via RC evaluate LLM's understanding on \emph{affirmative statements}, statements that describe a positive state in
the world. In this work, we focus on \emph{non-affirmative statements}, including \emph{negated statements}, and \emph{hypothetical statements} expressed via \emph{modals} and \emph{conditionals} (Section \ref{sec:hypotheticals}). When asked an affirmative question ("Who is tall?") over a negated context ("John is not tall") or a hypothetical  context ("If John was tall, he would have been a basketball player"), the  correct behavior is to indicate that the text does not answer the question. This ability to abstain is crucial for text-grounded tasks, as failing to abstain will result in an incorrect answer, which may lead to invalid claim verification or incorrect fact extraction.

Using imaginary instances,
we evaluate LLMs on non-affirmative constructions (Section~\ref{sec:hypotheticals}), and show that they often fail to understand these contexts, which is evident in their tendency {\em not} to abstain, and to provide an un-grounded answer. 

Next, we use the non-affirmative constructions to stress-test the context-faithfulness property (Section~\ref{sec:everything}). We show that in the case of non-affirmative contexts, LLMs are highly susceptible to knowledge conflicts.  When answering the questions on such contexts, they resort to consulting their parametric knowledge rather than relying exclusively on the text. This holds also for GPT-4, LLaMA 2 and Mixtral, which are proven to be context-faithful on affirmative contexts, but, as we show in Section~\ref{sec:everything}, this is  not so on  hypothetical ones.

To summarize, in this work we show that:
\begin{itemize}
\item When evaluating text understanding, we must neutralize the effect of parametric knowledge. This can be done using \emph{imaginary instances}.
\item Evaluating language understanding on such imaginary instances, we show that LLMs often fail to correctly understand non-affirmative statements, and in particular, statements that involve hypothetical situations (other ``possible worlds''). This is manifested by failing to abstain from answering questions  that are  unanswerable based on the text alone. 
\item When evaluating context-faithfulness,   it is not enough to rely on affirmative statements.  Assessing context-faithfulness on non-affirmative statements we show  that LLMs --- including ones that seem faithful on affirmative contexts --- are affected by their parametric knowledge when producing an answer.
\end{itemize}

All in all we observe that  the more involved the semantic phenomenon, the more susceptible the model is to knowledge conflicts influence. Hence, in the quest for trustworthy systems, further work should be devoted to both the {\em text-understanding} and {\em text-faithfulness} aspects, and more crucially, to their interaction. We hope that our benchmarks and evaluation framework will encourage further work and  improvements on these fundamental aspects of NLU.

\section{Overall Setup}

\paragraph{Extractive QA.} We work in an extractive question answering setup (also called {\em machine reading comprehenssion}, MRC) inspired by SQuAD2.0 \cite{rajpurkar2018know}, where the system is presented with a question and a context, and should provide an answer based on the context, or abstain from answering if the context does not answer the question (we instruct the model to return ``None'' in such cases). The questions are extractive, in the sense that for answerable questions, the answer  appears as a space-delimited span in the context.

\paragraph{Zero-shot.} We choose a zero-shot setup rather than in-context learning, for several reasons.
First, we believe that zero-shot, out-of-the-box, capabilities are a true indicator for determining whether an LLM is genuinely a general-purpose system \citep{qin2023chatgpt}. This is especially true for ``basic'' behaviors such as QA which are expected to work out-of-the-box.
Second, following this expectation, zero-shot QA aligns with interactions typical of end-users \citep{zhang2023language, kim-etal-2023-chatgpt, Zhou2024RelyingOT}. In real-world applications, the aim is not necessarily to directly manage a specific semantic phenomenon in question, but rather to generally handle language semantics accurately as a means to an end. Hence, the zero-shot capability becomes crucial.
Finally, the non-affirmative patterns we test (Section \ref{sec:hypotheticals}) are, by design, simple, and exhibit a consistent structure. This makes them easily learnable from examples. Therefore, providing them as in-context examples will likely result in the model replicating surface patterns, rather than demonstrating  ``understanding''.

\paragraph{Models and Prompts.} \label{models_and_prompts} We evaluate  \textbf{GPT-3.5} in two versions (turbo-0301 and turbo-0613), as well as \textbf{GPT-4} (0613), \textbf{LLaMA 2} (70B Chat) \citep{touvron2023LLaMA} and \textbf{Mistral} (Mixtral 8x7B Instruct v0.1) \citep{jiang2024mixtral}. We access the GPT models through OpenAI's API,\footnote{\url{https://openai.com/product}} using the default parameters, and LLAMA2 and Mistral through the Together AI API,\footnote{\url{https://api.together.xyz}} with a temperature of 0 and top-p=1. We experiment with several prompt templates for the QA task, and select the best templates for each model by optimizing the performance over 50 answerable and 50 unanswerable questions from SQuAD2.0 \cite{rajpurkar2018know}.\footnote{For the GPT models, the two best prompts performed similarly to each other, therefore we report results on both. For Mistral and LLAMA2, the best prompt was significantly better than all others, so we report results only on the best prompt.
} The prompt templates we use are available in  Appendix \ref{sec:basic}.

\paragraph{Simple Contexts.} We deliberately work with \emph{very simple contexts}, in order to demonstrate the effects of semantic contexts and knowledge conflicts 
in the cleanest possible settings. If the model fails to answer correctly, or fails to ignore parametric knowledge, when presented with a bare-bones, simple context, it is reasonable to assume it will fail also on more elaborate ones. Conversely, the simple context allows us to isolate the effect and attribute it to the modification we performed, without potential interferences of other properties of the text such as complex reasoning forms, special expertise requirements, and so on.

\section{Background}

\begin{figure*}[t!]
\centering
\includegraphics[width=0.98\textwidth]{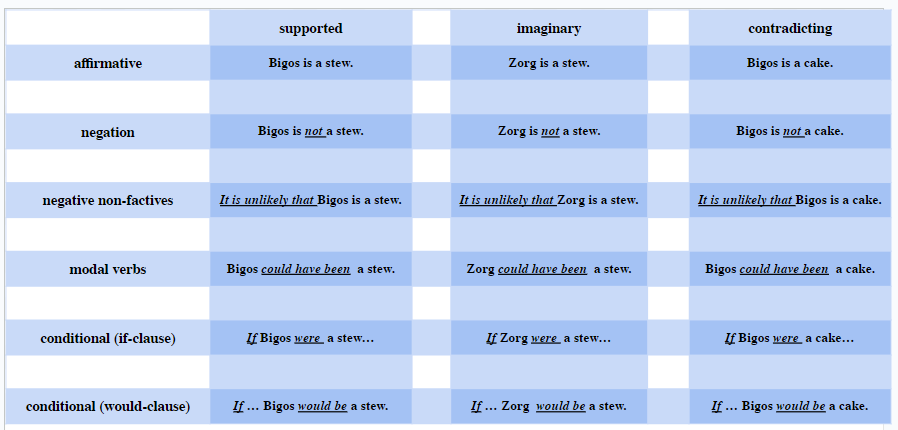}
\caption{The different variations we consider. Each column is a different relation with parametric knowledge, while each row is a different semantic modification.}
\label{fig:benchmarks}
\end{figure*}

\subsection{Parametric Knowledge Interferes with Reading Comprehension Assessment}

Machine Reading comprehension (MRC), implemented as answering questions over a textual context, has been considered a primary means to evaluate how well computer systems understand human language \cite{Lehnert78-book, hirschman-etal-1999-deep, chen2018neural,khashabi2019reasoningdriven, baradaran2020survey, sugawara-etal-2021-benchmarking}.
This approach, however, overlooks an important problem, typical of contemporary LLMs, which hinders its effectiveness: the LLM's parametric knowledge acquired in pre-training might clash with the textual context provided in the prompt.

This problem, known as a \emph{knowledge conflict}, has been addressed in an extensive body of recent work \citeg{DBLP:journals/corr/abs-2109-05052,neeman2022disentqa,li2022large,zhou2023contextfaithful,Xie2023AdaptiveCO,mallen2023trust,qian2023merge}, mostly in question-answering (QA) settings, covering diverse models and scenarios.
These works show that knowledge conflicts distort language models' behavior, leading to numerous erroneous trends, such as answering questions based on the internal, parametric rather than externally provided contextual knowledge, hallucinations, or abstention where a correct answer  based on the context is expected (e.g. prompt: \emph{"Context: Elon Musk is a business magnate and investor. He is the owner and CEO of Twitter. Question: Who is the CEO of Twitter?"}, model's answer: \emph{"Jack Dorsey"}. Example by \citet{zhou2023contextfaithful}).

In this work we approach knowledge conflicts from an orthogonal direction. That is, instead of simply claiming that knowledge conflicts harm NLU capacities, we underscore that knowledge conflicts hinder {\em our ability to assess} LLMs' NLU capabilities through the RC task. This is because the results of such an assessment might reflect a \emph{mixture} of models' linguistic capabilities and knowledge-conflict effects, that are hard or impossible to discern. Hence, as a first step we suggest an approach that allows us to disentangle these two factors, and assess current LLMs' language understanding abilities without interference of knowledge-conflict effects.

\subsection{The Importance of Knowing  to Abstain from Answering in Text-Grounded Tasks}

It has been observed many times that LLMs often confidently provide incorrect responses or answer questions without a definitive solution, potentially leading to hallucinations \citep{kasai2024realtime, liu2024examining, deng2024gotcha}.  As \citet{mik2024caveat} puts it, ``The dangerous combination of fluency and superficial plausibility leads to the temptation to trust the generated text and creates the risk of overreliance". Therefore, the ability of LLMs to abstain instead of producing incorrect answers is crucial for trustworthy AI \citep{rajpurkar2018know, neeman2022disentqa, zhou2023contextfaithful, chen2023adaptation, varshney-baral-2023-post, feng2024dont}. While in certain scenarios, abstention might be expected in cases of knowledge conflicts \citep{feng2024dont} or below a certain confidence threshold \citep{DBLP:journals/corr/abs-2109-05052}, in this work (similarly to \citet{zhou2023contextfaithful}) we focus on the models' ability to refrain from answering when, \textit{based on the context alone}, the  answer is simply absent.

Such abstention ability is crucial for prompt-grounded cases such as retrieval augmented generation (RAG) \citep{shi2023trusting, mallen2023trust, Zhang2023MergingGA, asai2023selfrag} or claim verification \citep{10.1162/tacl_a_00454,10.1162/tacl_a_00486, wang2023explainable}, where we attempt to verify an existing claim based on some text. A model's failure to abstain is beyond a minor nuance, as it could result in an inaccurate verification. Moreover, answering based on parametric knowledge rather than abstaining may  lead to a wrong verification in cases the parametric knowledge about the topic is incorrect.

The ability to abstain becomes increasingly important in the context of hypothetical constructions, as we elaborate here in turn.

\subsection{
Understanding Hypothetical Scenarios: Modality and Conditionals}
So far, evaluating LLMs' NLU capacity in the context of RC, and in particular the ability to abstain, has been  done on    {\em affirmative} statements, that is, statements that describe factual state of affairs. However, from a predicate-argument structure point of view, these are rather straightforward constructions. 
Formal semanticists  investigate the influence of operators such as {\em negation}, {\em conditionals} or {\em modals} on the interpretation of the text. Likewise, we aim to advance the study of  {\em natural language understanding} in LLMs by investigating how they cope with these more complex semantic constructions.
In this subsection 
we provide a brief overview of modal and conditional constructions expressing \textit{hypothetical} situations, which are central to our work.

 \textbf{Modality} expresses the relation of the utterance to reality
\citep{khomutova}. Modal statements do not describe how things actually are, but rather convey plans, desires, norms (\emph{deontic} modality), or plausibility (\emph{epistemic} modality) (there are also more fine-grained taxonomies of modal senses).
Modality can be conveyed through a diverse, virtually open-ended, range of linguistic expressions \cite{pyatkin-etal-2021-possible}.
In this study we focus on \emph{epistemic} modality expressed through modal verbs (\emph{could, might, may}) and negative non-factives (\emph{It is unlikely that..., it is impossible that...} etc.).

 \textbf{Conditionals} convey the dependence of one situation on another, often involving an ``if-then'' clause. In English, the most common conditional constructions are \emph{real conditionals} expressing habitual or possible situations (\emph{If I'm tired, I rest; If she has time, she'll come}) and unreal conditionals describing situations that are unlikely, untrue or were untrue in the past (\emph{If I were sick, I would stay home; If you had asked, she would have told you the truth}). In this work we focus on \emph{unreal} conditionals.

Both modality and conditionals denote hypothetical, alternative scenarios, sometimes referred to as other \emph{possible worlds}, as in  \citet{kripke}.\footnote{In fact, conditional mood (e.g., English would) is often regarded as an expression of modality \cite{stephany_1986}.} 

This ability to think in terms of alternative realities, known as \emph{modal displacement}, is ``unique to human language, universal to all languages, and acquired early'' \citep{matthewson_2016}. By contrast, non-modal affirmative and negative statements focus on a single scenario leaving any alternatives aside.
In this work we specifically explore whether this contrast is challenging for contemporary LLMs' comprehension capabilities.

\begin{figure*}[t!]
\includegraphics[width=\textwidth]{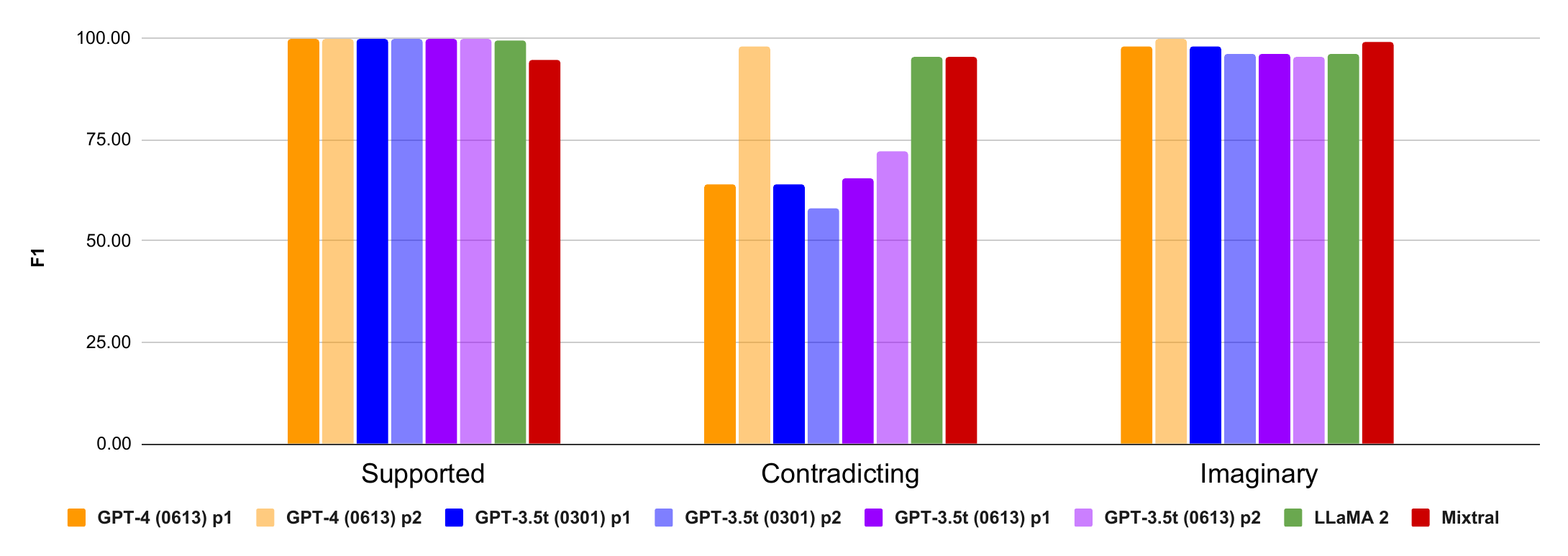}
\caption{F1 scores on the affirmative versions of the \emph{supported}, \emph{contradicting} and \emph{imaginary} sets.}
\label{fig:affirmatives}
\end{figure*}

\section{Evaluation through Imaginary Worlds} 
\label{method}

Parametric knowledge may distort text understanding evaluation results. When the answer is supported by the parametric knowledge, the model may answer correctly without consulting the text. Conversely, when the text conflicts with the parametric knowledge, the model may answer (incorrectly) based on the parametric knowledge despite understanding the text. 

We thus stress that in order to reliably evaluate text understanding through QA, we must use data that \emph{is neither supported by nor contradicts} the model's internal parametric knowledge. This can be done by creating {\em imaginary} reading-comprehension instances. We create such data based on existing QA benchmarks, by substituting all real-world entities and events in questions and contexts with fictitious, \textit{imaginary} ones.
E.g., in the factual context-question pair \textit{Bigos is a stew -- What is Bigos?}, \textit{Bigos} is replaced with an imaginary entity \textit{Zorg}. This results in an {\em imaginary} QA pair: \textit{Zorg is a stew. -- What is Zorg?}.

\subsection{Validating Imaginary Worlds}\label{imaginary}
We validate the effectiveness of using imaginary worlds by comparing them to \emph{counterfactual worlds}, suggested in previous work  \citep{DBLP:journals/corr/abs-2109-05052,neeman2022disentqa,li2022large,zhou2023contextfaithful,Xie2023AdaptiveCO,DBLP:conf/emnlp/ChenZC22},
in which the context is replaced with one that contradicts the parametric knowledge.

\paragraph{Data Creation}
We take 50 questions from the WebQuestions dataset \cite{berant-etal-2013-semantic} that contains linguistically simple questions about facts 
(e.g. \emph{What did Mary Wollstonecraft fight for?}), for which ChatGPT provides a correct parametric answer. These 50 questions, together with their ChatGPT answers, are taken to be our \emph{supported set}, where the ChatGPT answer serves as the context for each question:

\begin{quote} 
\textbf{Context}: Mary Wollstonecraft fought for women's rights.

\textbf{Q}: What did Mary Wollstonecraft fight for?

\textbf{Possible answers}: Mary Wollstonecraft fought for women's rights; Wollstonecraft fought for women's rights; for women's rights; women's rights
\end{quote}

From these, we derive \emph{contradicting} (counterfactual) and \emph{imaginary} sets: 

\noindent The \textbf{contradicting} data is created by replacing the factual answer spans in the contexts with other, \emph{counterfactual}, ones (while the question remains the same):
\begin{quote} 
\textbf{Context}: Mary Wollstonecraft fought for \emph{immigrants'}s rights.

\textbf{Q}: What did Mary Wollstonecraft fight for?

\end{quote}

\noindent The \textbf{imaginary} data is created by replacing the entities in both the context and the questions with made-up, imaginary ones:\footnote{The modifications to obtain imaginary and contradicting data are obtained with ChatGPT's assistance, but each example is reviewed and edited manually.}

\begin{quote} 
\textbf{Context}: \emph{The Zogloxians} fought for women's rights.

\textbf{Q}: What did \emph{The Zogloxians} fight for?

\end{quote}

To summarize, we create three sets, each consisting of 50 context-question-answer triplets: supported, contradicting, and imaginary, where the imaginary set does not intersect with models' world knowledge, eliminating knowledge-conflict effects. 

In all sets, the context trivially answers the questions, and we expect the language models to succeed in answering correctly.
\paragraph{Results} \label{affirmative_results} Figure \ref{fig:affirmatives} shows the LLM's performance on the three sets, measured by SQuAD F1 (see Appendix \ref{sec:eval} for evaluation details). On the \emph{supported} set all models achieve near-perfect or perfect scores. However, we cannot tell if the answers come from the text or from the parametric knowledge, i.e., the results might be inflated by reliance on the models' inner knowledge. In contrast, on the \emph{contradicting} set, the first GPT-4 prompt and both GPT-3.5 prompts obtain F1 below 70\%, as the conflict with their parametric knowledge causes them to output the parametric answer or incorrectly abstain on many instances. Finally, on the \emph{imaginary} set, all models perform near perfectly, but slightly inferior to the \emph{supported} set. 
This suggests that only the performance on imaginary data reliably showcases LLMs' NLU abilities, free from distortions caused by knowledge conflict or alignment effects, validating the importance of imaginary-worlds approach for accurate language understanding evaluation.

\section{Understanding Non-Affirmative Texts}\label{sec:hypotheticals}

\begin{figure*}[t!]
\includegraphics[width=\textwidth]{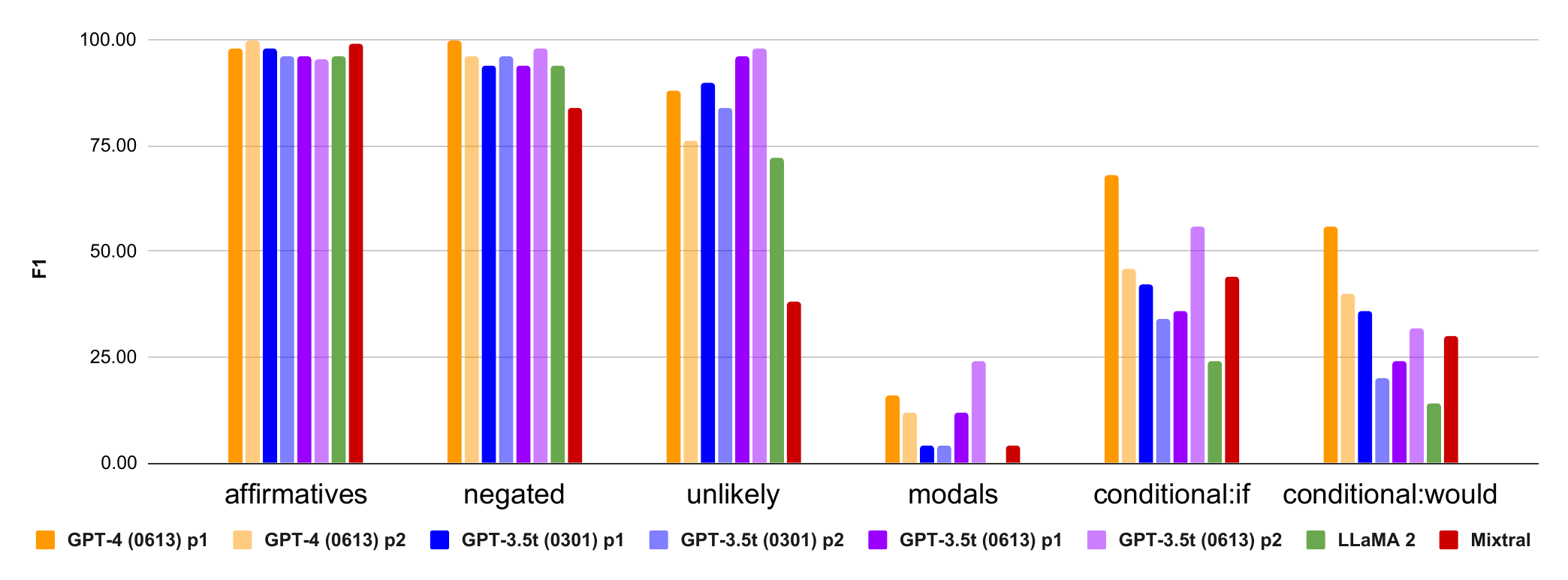}
\caption{Comparing performance over different semantic variations using the Imaginary setting.}
\label{fig:imaginary}
\end{figure*}

How well do LLMs understand non-affirmative constructions? To assess this, we derive non-affirmative variations of the \emph{imaginary} set. Each variant changes the context from its affirmative form to a non-affirmative one. Figure \ref{fig:benchmarks} summarizes the different conditions.

\paragraph{Negative constructions}
\begin{quote}
 
 {\em Negation} (The Zogloxians \underline{did not} fight for women's rights).
 
 {\em Negative non-factives}, i.e. non-factive predicates expressing strong doubt, such as \emph{unlikely/doubtful/improbable} (\underline{It is unlikely that} the Zogloxians fought for women's rights).
 \end{quote}
 
\paragraph{Hypothetical constructions} (see Section \ref{sec:hypotheticals})
\begin{quote}
 {\em Modal verbs} (The Zogloxians \underline{might have} fought for women's rights).
 
 {\em Conditionals 1:} Transforming the context into an if-clause of unreal conditional (\underline{If} the Zogloxians had fought for women's rights, they might have contributed to gender equality in their society.).
 
 {\em Conditionals 2:} Transforming context into the would-clause of unreal conditional (If they were part of a progressive society, the Zogloxians \underline{would have} fought for women's rights).
 
\end{quote}

Thus, we expand the imaginary benchmark with 5 additional test sets, each comprising 50 examples and representing a distinct semantic phenomenon.
The modifications are obtained with ChatGPT's assistance, but each example is reviewed and edited manually.

Following the modifications, all  questions on these instances \textbf{cannot be answered} with a context span, because the modifications cancel the entailment between the context and the answer. Thus, \textbf{the models are expected to abstain.}\footnote{Crucially, we intentionally alter only the \textit{contexts}, leaving the questions unchanged. This is because modifying the questions accordingly (e.g., \textit{"What might the Zogloxians fight for?"}) would render them answerable once again, whereas our goal is to make the answer uninferable from the context.}

\subsection{Results} Figure \ref{fig:imaginary} shows the results for all models.

\paragraph{Quantitatively} With the exception of Mixtral, the models are quite robust at recognizing simple negation (though, somewhat surprisingly, they do suffer some degradation relative to simple affirmatives). However, when we move to negative non-factives (``unlikely''), most models -- especially LLaMA 2 and Mixtral -- exhibit a large decrease in performance. On the hypothetical contexts (modals and conditionals) we observe a very dramatic drop for all models, with modal verb semantics being particularly challenging to understand.

\paragraph{Qualitatively}
Manual error analysis of ChatGPT's responses reveals that for the non-affirmative contexts, where the model should abstain, 94.64\% of the errors involve ignoring entailment-cancelling modifications (e.g., negation, modality) and answering as if the context is affirmative. For example, in the 
Context: \emph{“If they were part of a progressive society, the Zogloxians \underline{would have fought} for women's rights.”} Q: \emph{“What did the Zogloxians fight for?”} The answer is: \emph{“Women's rights”}. (That is, the model essentially discards the conditional.)
\subsection{Discussion}

\paragraph{Modal displacement challenge.} Thus, all the tested LLMs struggle with \textit{modal} and \textit{conditional} contexts whose understanding requires \textit{modal displacement}, ability to think in terms of alternative realities.
Facing modal and conditional semantic modifications, the models often overlook them, behaving as if the statement is affirmative.

\paragraph{Non-factives' mixed semantics.} Contexts with \textit{negative non-factives} (e.g. \textit{\underline{It is unlikely} that Zorg is a stew}) stand out in this picture: ChatGPT (especially with prompt 2) handles them accurately, along with affirmatives and negatives,  whereas the performance of other LLMs'  on these contexts noticeably drops --- though not as dramatically as on other hypothetical contexts. 

This is likely due to the fact that negative non-factives (e.g., \textit{unlikely, impossible, improbable}) combine modality and negation in their semantics. So ChatGPT seems to treat them as negatives, while other models interpret them as modals or as a ``transitional form'' between negatives and modals.
\section{Re-assessing Context-Faithfulness}\label{sec:everything}

\begin{figure*}[t!]
\includegraphics[width=\textwidth]{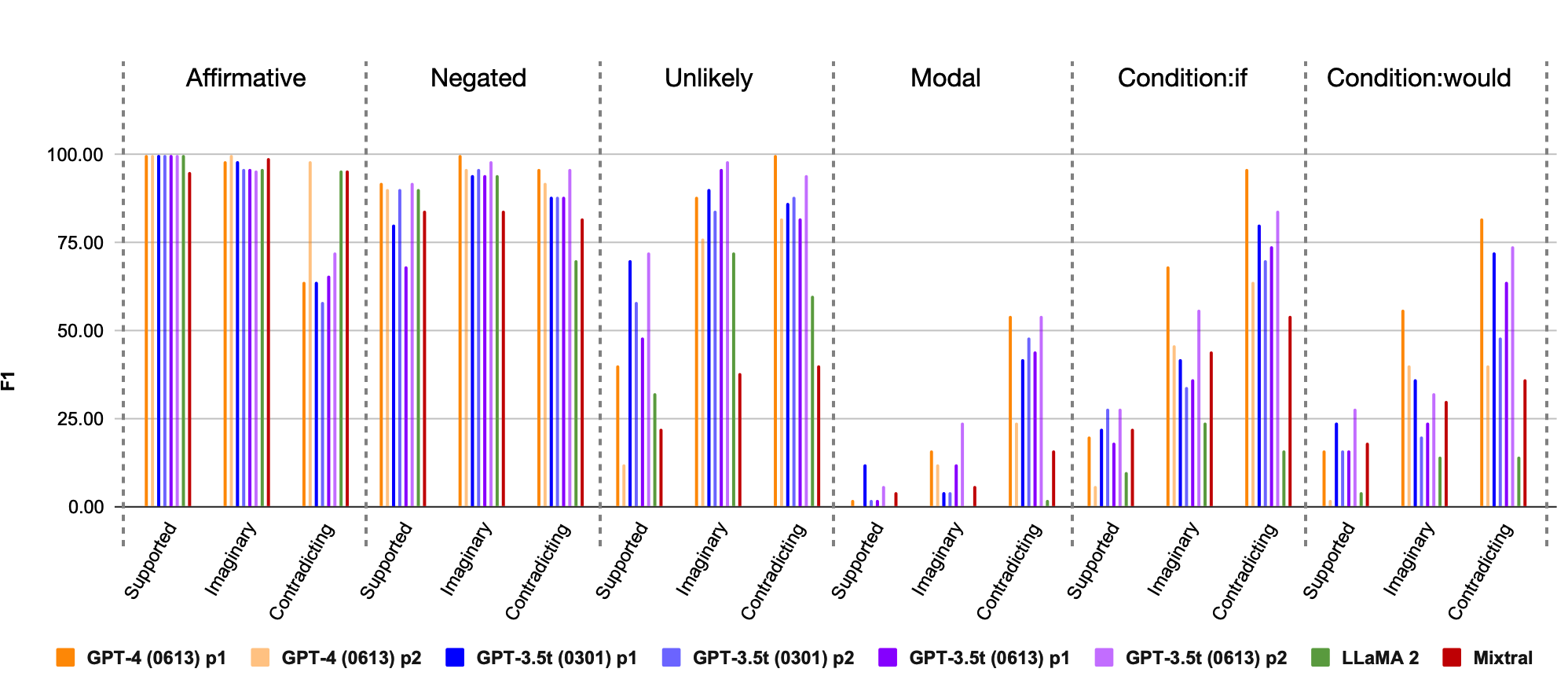}
\caption{Comparing the different knowledge-conditions (supported, contradicting, imaginary) across different semantic conditions.}
\label{fig:everything}
\end{figure*}

Having identified that the LLMs struggle with understanding non-affirmative statements, and in particular hypothetical data, we are now set to re-assess their context-faithfulness, this time using the non-affirmative constructions on the \emph{supported} and \emph{contradicting} sets, and comparing the results to the \emph{imaginary} set.  

Recall that in all the non-affirmative cases, \textbf{the context does not answer the question}, and the model should abstain. However, the non-affirmative \emph{supported} cases are derived from affirmative statements consistent with the models' parametric knowledge; the non-affirmative \emph{contradicting} cases are derived from affirmative statements that contradict the parametric knowledge. Nevertheless, a context-faithful model would ignore the parametric knowledge and respond as in the \emph{imaginary} case. By contrast, a model that fails to ignore its parametric knowledge, will struggle to abstain on the \emph{supported} set, scoring lower compared to the imaginary set. On the other hand, it will abstain more on the \emph{contradicting} set, scoring higher than on the imaginary set. Thus, the comparison to the imaginary data results allows us to gauge the models' context-faithfulness.

\subsection{Results} Figure \ref{fig:everything} shows the results. 

\paragraph{Quantitatively} Overall, we observe significant performance differences between the three sets, imaginary, supported and contradicting, which follow the expected trend: for supported sets the overall scores on non-affirmatives are lower than for imaginary sets, and for contradicting sets -- higher. This confirms the LLMs' susceptibility to knowledge conflict effects. The effect is consistent across all models and prompts, except for LLaMA 2, where the contradicting set results are not higher, and in some cases lower, than the imaginary set scores.
For all the models, the more challenging {\em modals} and {\em conditionals} exhibit the greatest performance gaps between the benchmarks, i.e., increased vulnerability to knowledge conflicts. Below we analyze more nuanced trends for each specific data group.

\textit{Affirmative contexts}: for most models the performance is best on supported data (\textit{aligned} with the LLM's inner knowledge) and worst on contradicting data (\textit{conflicting} with the model's inherent knowledge). However, GPT-4 (with prompt 2), LLaMA 2 and Mixtral are practically unaffected by parametric knowledge and knowledge conflicts on affirmative and negative contexts, seemingly marking a notable advance in context-faithfulness.

\textit{Negative contexts}: most models handle negative contexts quite accurately, with minimal performance differences between  imaginary, supported and contradicting benchmarks. Across all models, performance is highest for imaginary contexts and generally decreases for both supported and contradicting data. OpenAI LLMs show the poorest performance on supported data, while LLaMA and Mixtral perform worst on contradicting data.

\textit{Hypotheticals (modals and conditionals)}, demonstrate the weakest performance on supported data, and most models --- except LLaMA 2 --- perform best on contradicting data. For LLaMA 2, the performance on contradicting data is either similar to or worse than on imaginary data.

\textit{Negative non-factives (``unlikely'')}, 
preserve the intermediate status between hypotheticals and negatives observed in Section \ref{sec:hypotheticals}: ChatGPT and LLaMA 2 handle them like negatives (i.e. perform best on imaginary data). GPT-4 and Mixtral handle them like hypotheticals, with best performance on contradicting data.

\paragraph{Qualitatively} We perform manual error analysis of 2880
responses obtained on the supported and contradicting data: 480 for each semantic variation (affirmative, negated, etc.)
This manual error analysis helps us explain what error types stand behind the performance differences between supported, imaginary and contradicting data.

\textit{Affirmative contexts }(context-span answer expected):
 The vast majority of errors consist in abstention.
 Another error type, specific to the contradicting data, involves answers based on the LLM's world knowledge rather than the context. E.g.: Context: \emph{“The Deutsche Mark is the currency of Germany now.”} Q: {”What is the currency of Germany now?”} Prediction: {”The Euro.”} The answer is factually correct, but contradicts the context. Both error types are typical of contradicting data (accounting for 93\% and 7\% of the errors respectively) explaining the performance drop on this data type.

 \textit{Non-affirmative contexts} (abstention expected):
The most frequent error types are as follows:
 \emph{(i) Ignoring the entailment-cancelling modifications} (such as negation, modality etc.) and behaving as if the context is affirmative. E,g.:
Context: \emph{“If Ryan Reynolds' romantic choices had been different, in 2012, he \underline{would have been} married to Scarlett Johansson.”} Q: \emph{“Who was Ryan Reynolds married to in 2012?”} Prediction: \emph{“Scarlett Johansson”}. (Here the model ignores the conditional.) This is the most frequent error type accounting for about half (48.25\%) of  the mistakes across all non-affirmative contexts.
 (ii) Answers based on world knowledge rather than on context. These are only found in the contradicting data.
Context: \emph{The Deutsche Mark may be the currency of Germany now.} Q: \emph{What is the currency of Germany now?} Prediction: \emph{Euro}. The answer is factually correct (in our world at this specific moment),
but based only on world knowledge, as the text provides no hint for such an answer. Such mistakes are  responsible for 22\% of all the errors on the contradicting benchmarks. Importantly, such mistakes are the most frequent for negatives (69.0\%) and non-factives (16.22\%).

Manual analysis of outputs for both affirmative and non-affirmative contexts also suggests that \textit{contradicting} contexts \textit{trigger abstention}. This explains the drop in performance on \textit{affirmative} contexts (where a context span answer is expected) and the higher accuracy on \textit{non-affirmative} contradicting contexts (where abstention is expected), suggesting that on the latter the LLMs are frequently ``right for the wrong reasons''.

This also explains the difference in patterns between modals and conditionals on one hand and negatives and non-factives on the other. Contradicting contexts are associated with two main behaviors: \textit{abstention} and \textit{parametric outputs}.
Negation (and for some models also non-factives) trigger more parametric answers, which are incorrect, and result in lower scores on the contradicting set. On the other hand, modals and conditionals trigger more abstention, which is the correct behavior (but for the wrong reason), leading to relatively higher scores on the contradicting set.

Now the last question remains. Why is there no similar performance improvement  on contradicting modal and conditional data in LLaMA 2? Manual inspection of LLaMA 2 outputs shows that tis model exhibits a different reaction to knowledge conflict: the tendency to output parametric answers here is stronger than in other models (30\% of all the errors), rather than abstention. This leads to a \textit{drop} in performance on contradicting data. Hence, \textbf{the contrasts in quantitative trends between LLaMA 2 and other LLMs stem from divergent responses to knowledge conflicts} --- LLaMA 2 often provides a parametric answer rather than abstaining --- and not from genuine differences in the LLMs' abilities.

\subsection{Discussion}
\paragraph{Summarizing Overall Trends}\label{trends}

Susceptibility to knowledge conflict/alignment effects is evident across all the models.

Semantic difficulty correlates with stronger knowledge conflict effects: \textit{hypothetical} (modal and conditional) contexts, which proved more challenging in the NLU tests (see Section \ref{sec:hypotheticals}), also exhibit more pronounced gaps between the supported, imaginary, and contradicting benchmarks (as is obvious from Figure \ref{fig:everything}).

For some models, the knowledge conflict effects are practically absent when handling affirmative  (for GPT-4 with prompt 2, Mixtral, LLaMA 2) and negative  (for GPT-4 with prompt 2, and Mixtral) contexts, \emph{suggesting that these models may be context-faithful}. However, the issue resurfaces when the same models deal with hypothetical contexts, revealing that they are still text-unfaithful.

\textit{Contradicting} contexts w.r.t to parametric knowledge typically \textit{trigger abstention}. This behavior is consistent across affirmative and non-affirmative constructions. This leads to higher accuracy on non-affirmative contexts (where abstention is expected), but to lower accuracy on affirmative contexts (where a context span answer is expected).

Conversely, \textit{supported} contexts (whether affirmative or not) trigger the tendency to output a context span answer (which in this case coincides with the parametric answer). In other words, if a text span is consistent with the parametric answer, the model is more likely to predict it even if the surrounding text cancels the statement. This leads to higher accuracy on affirmative contexts (where such answers are expected), but lower accuracy on modified contexts (where the model should abstain).

\paragraph{Characterizing the models' behavior} Our error analysis results allow us to sketch a working hypothesis about the LLMs' behavior when performing reading comprehension. If the LLM \textit{initially decides that the question is \textbf{unanswerable}} (in our experiments - mostly on negated contexts and partially on non-factives), it looks for a parametric answer. If it does not find one (as observed in imaginary data), it abstains; if found (in supported and contradicting data), it decides between abstaining and providing the parametric information. Conversely, \textit{when the LLM initially determines that the question is \textbf{answerable}} (whether accurately or by overlooking semantic modifications), it then \textit{compares} the contextual answer to the parametric one. If they \textit{coincide}, the LLM tends to output the answer; if they \textit{differ}, it triggers a strong tendency to abstain (except for LLaMA 2). This logic determines a number of more fine-grained trends (e.g. the error of outputting the parametric answer is especially typical of negative contexts).
That being said, this remains a hypothesis, and further evidence is required to definitively confirm it.

\section{Prompt-based Mitigation Experiments}
\label{additional}

\begin{figure}[t!]
\includegraphics[width=\linewidth]{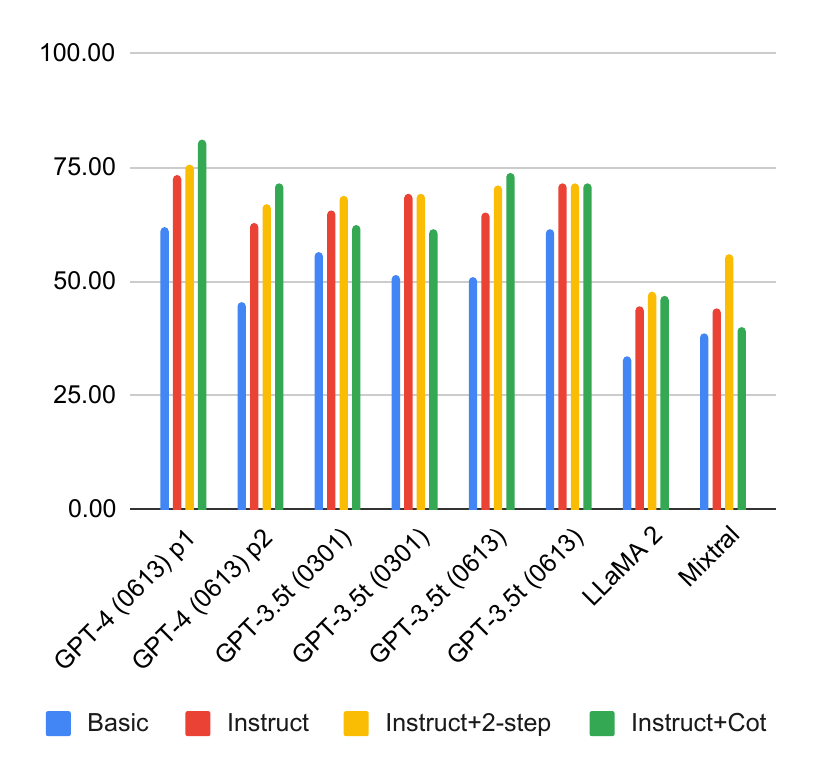}
\caption{Comparing prompting approaches. Each bar's height is the average score across all non-affirmative cases for this model and prompting technique.}
\label{prompt_comp_fixed}
\end{figure}

We experiment with additional prompting techniques: instructed prompting \citep{zhou2023contextfaithful}, two-step prompting \citeg{singhal2023expertlevel,lu2023hybrid} and chain-of-thought (CoT) \citep{kojima2023large} combined with instructed prompting.

For \emph{instructed prompt} experiments we prepend the \emph {basic prompts} (\S\ref{models_and_prompts}) with an \emph{explicit instruction to ignore the model's inherent world knowledge and base the answer solely on the context}.\footnote{For each combination of the model version and the basic prompt (prompt 1 and gpt-3.5-turbo-0301; prompt 2 and gpt-3.5-turbo-0301; prompt 1 and gpt-3.5-turbo-0613 etc.) we search for the right formulation of the instruction, 
testing them on the same 100 examples from SQuAD2.0 which we used for \emph{basic prompt} selection (\S\ref{models_and_prompts}) The exact instruction phrasings are available in Appendix \ref{sec:instr}.}

For \emph{two-step prompting} we use the \emph{instructed prompts}, and add a second step: asking the model if the context contains evidence for the predicted answer (unless the model predicts that the question is unanswerable). If the model answers “no”, we change the initial prediction to “unanswerable”.

For CoT prompting we combine the instructed prompt\footnote{For CoT experiments we once again search for the right formulation of the instruction for each combination of the model and the basic prompt. The exact instruction phrasings are available in Appendix \ref{sec:instr}.} and the line \textit{"Let's think step by step"} as  suggested in \citet{kojima2023large}.

We run the same set of experiments as described in Section 
\ref{sec:everything}, using the three prompting techniques mentioned above. 
Figure \ref{prompt_comp_fixed} shows the scores of the different techniques for each model, where each score is an average over all the non-affirmative cases (negation, unlikely, modals, conditional:if and conditional:would). 

First, while the prompting techniques overall result in substantial increase in accuracy, they still fall short of achieving perfect performance. Additionally, while not shown in the aggregate figure, \textbf{the \emph{trends} observed in the basic prompt experiments, persist across all the prompting approaches}. In particular, there is still a pronounced split in results between hypothetical (modal and conditional) and non-hypothetical (affirmative and negative) contexts. Just like before, we observe both an \textit{accuracy drop} and \textit{more intense knowledge conflict effects} on the hypothetical contexts, although the disparities between the imaginary, supported and contradicting sets become somewhat smaller, especially for GPT-4.

Looking into individual techniques, instructed prompting consistently enhances overall results across most models and setups.
For most models, two-step prompting further improve the results, suggesting  that in NLI setting (the second step in two-step prompting) LLMs might be able to handle hypotheticals better than in QA setting.

CoT prompting combined with instructed prompting showed promising results on the subset of 100 SQuAD instances previously used for prompt selection (\S\ref{models_and_prompts}), consistently outperforming instructed prompting alone across all models.\footnote{As opposed to other prompting techniques we tested on the same SQuAD subset, including system prompts, prompts with detailed reasoning steps, outputting results in JSON format, prompt chaining, self-consistency prompting, opinion-based prompting and combinations thereof.} However, when applied to our non-affirmative sets, this prompting method did not show consistent improvements: it enhances the results for certain models and data types (such as GPT-3.5 turbo-0613, GPT-4 and LLaMA 2), while negatively impacting others (GPT 3.5t 0613 and Mixtral).

To summarize, the use of certain prompting strategies can improve the overall accuracy and somewhat mitigate (but by no means eliminate) knowledge conflict effects. At the same time, the erroneous trends revealed in prior experiments, and in particular the limitations in understanding hypothetical --- modal and conditional --- scenarios, persist {\em regardless} of the prompting approach used.

\section{Relation to Previous Work}

\label{sec:prev_work}

 While LLMs' ability to independently produce impressive results using parametric knowledge is beneficial, issues arise in context-specific tasks as well as in retrieval-augmented generation (RAG). 
 Conflicts between contextual and parametric knowledge cause models to overlook vital contextual cues, resulting in errors. A growing body of work explores these knowledge-conflict issues.\footnote{See \citet{Shaier2024DesiderataFT} for an extensive survey of knowledge-conflict related work.} 

Our work differs from previous studies in several aspects. While most studies \citeg{DBLP:journals/corr/abs-2109-05052, neeman2022disentqa, li2022large, zhou2023contextfaithful, Xie2023AdaptiveCO} contrast factual and counterfactual knowledge  (or parametric knowledge with directly conflicting contextual knowledge), we introduce ``imaginary data'' that neither confirms nor contradicts the model's internal knowledge. It should be noted that, while our approach was developed independently, similar ideas have been explored in earlier reading comprehension literature. \citet{kočiský2017narrativeqa} substitute named entities with markers ``to build representations for entities from stories that were never seen in training''. Their idea was, in its turn, inspired by \citet{hermann2015teaching} who replace entities with abstract entity markers to build a corpus for ``evaluating a model's ability to read and comprehend a single document, not world knowledge or co-occurrence''. However, to our knowledge, we are the first ones to use imaginary data in the context of knowledge conflict in current LLMs.

 While most previous studies use longer, often multiple, contexts \citeg{Xie2023AdaptiveCO,DBLP:conf/emnlp/ChenZC22,mallen2023trust,Jin2024TugofWarBK}, we intentionally focus  on single, concise, linguistically straightforward contexts. This approach isolates the influence of knowledge conflict  on semantic understanding {\em without} interference of confounding factors such as complex reasoning, special expertise etc.
 
Moreover, while many studies experiment with irrelevant contexts in order to see how LLMs handle them
\citeg{neeman2022disentqa,li2022large,zhou2023contextfaithful,qian2023merge}, to the best of our knowledge, our study is the first to explore irrelevant contexts with parametric answer strings serving as distractors for the model (see experiments with modified supported data in Section \ref{sec:everything}).

More importantly, our work prioritizes \textit{language understanding} over knowledge conflict as such. Our work is the only one, to the best of our knowledge, that extensively explores the effect of semantic modifications added to both supported and unsupported (i.e. imaginary and contradicting) contexts. We highlight knowledge conflicts as an essential consideration rather than the primary focus and extend the inquiry beyond a mere restatement of a known problem by exploring the realm of ``possible worlds'' introduced by hypothetical constructions, and show how knowledge conflict undermines the comprehension  of  hypothetical statements. Crucially, while related studies typically assume that the model understands the context correctly (with some exceptions in \citet{DBLP:conf/emnlp/ChenZC22}),
our study 
clearly demonstrates  accurate as well as \textit{erroneous} context interpretation by the model. 

Many recent studies highlight LLMs' overreliance on prior knowledge as a major source of hallucination \citep{shuster-etal-2021-retrieval-augmentation, Ji_2023, Xie2023AdaptiveCO} and explore RAG as a means to reduce hallucination and improve factuality in LLMs \citeg{shi2023trusting, mallen2023trust, Zhang2023MergingGA, asai2023selfrag}. However, in order for RAG to be effective, LLMs need to correctly determine the relevance of external contexts and remain context-faithful when using the retrieved evidence. Our study highlights the critical significance of basic language understanding skills for LLMs to generate truly text-grounded output.

To sum up the space of related work in the realm of knowledge conflicts, Table \ref{rel} (Appendix \ref{sec: rel_appendix}) compares our study with related research across various characteristics.

\section{Conclusions}
In this work we aim to assess the language understanding capabilities of contemporary LLMs for {\em text-grounded prompting}, in light of two different perspectives: the {\em semantic} perspective, that is, how well do models understand certain complex constructions, and the {\em epistemic} perspective, that is, how are NLU capabilities of LLMs  affected by the LLM's internal knowledge, whether supporting or contradicting the context provided in the prompt. We explore six linguistic environments (affirmative, negated, negative non-factive, modal, conditional:if and conditional:would) across three types of data instances: supported, contradicting, and neutral ("imaginary"). Our experiments clearly show that, not only do contemporary LLMs struggle with the semantics of ``possible worlds'', or alternative realities, expressed by modals and conditionals, but also these semantic constructions trigger knowledge-conflict effects: the failure to  correctly understand such contexts  causes models to resort again to their parametric knowledge, rather than remaining faithful to the text. We underscore three take-home messages for the NLU research community. First, that faithful evaluation of language understanding abilities can and should be done on imaginary data,  avoiding epistemic clashes between the model parametric knowledge and textual context. Second, that any NLU investigation done on non-imaginary (supported or contradicting) contexts, should strive to disentangle the mixed effect of text-understanding and text-faithfulness, prior to making any bold statements about either. And finally, despite efforts to the contrary, current LLMs are not context-faithful, and often prefer their parametric knowledge over the provided text. For LLM users, this suggest being careful when using text-guided prompting in sensitive verification scenarios. For LLM developers, this suggests looking beyond the simple, affirmative contexts when developing context-faithfulness abilities.

\section*{Acknowledgements}
This work has been funded by the Israel Science Foundation, grant number 23/670.

\bibliography{tacl2021}

\begin{thebibliography}{51}
\expandafter\ifx\csname natexlab\endcsname\relax\def\natexlab#1{#1}\fi

\bibitem[{Asai et~al.(2023)Asai, Wu, Wang, Sil, and Hajishirzi}]{asai2023selfrag}
Akari Asai, Zeqiu Wu, Yizhong Wang, Avirup Sil, and Hannaneh Hajishirzi. 2023.
\newblock \href {http://arxiv.org/abs/2310.11511} {Self-rag: Learning to retrieve, generate, and critique through self-reflection}.

\bibitem[{Atanasova et~al.(2022)Atanasova, Simonsen, Lioma, and Augenstein}]{10.1162/tacl_a_00486}
Pepa Atanasova, Jakob~Grue Simonsen, Christina Lioma, and Isabelle Augenstein. 2022.
\newblock \href {https://doi.org/10.1162/tacl_a_00486} {{Fact Checking with Insufficient Evidence}}.
\newblock \emph{Transactions of the Association for Computational Linguistics}, 10:746--763.

\bibitem[{Baradaran et~al.(2020)Baradaran, Ghiasi, and Amirkhani}]{baradaran2020survey}
Razieh Baradaran, Razieh Ghiasi, and Hossein Amirkhani. 2020.
\newblock \href {http://arxiv.org/abs/2001.01582} {A survey on machine reading comprehension systems}.

\bibitem[{Berant et~al.(2013)Berant, Chou, Frostig, and Liang}]{berant-etal-2013-semantic}
Jonathan Berant, Andrew Chou, Roy Frostig, and Percy Liang. 2013.
\newblock \href {https://aclanthology.org/D13-1160} {Semantic parsing on {F}reebase from question-answer pairs}.
\newblock In \emph{Proceedings of the 2013 Conference on Empirical Methods in Natural Language Processing}, pages 1533--1544, Seattle, Washington, USA. Association for Computational Linguistics.

\bibitem[{Chen(2018)}]{chen2018neural}
Danqi Chen. 2018.
\newblock \emph{Neural Reading Comprehension and Beyond}.
\newblock Ph.D. thesis, Stanford University.

\bibitem[{Chen et~al.(2022)Chen, Zhang, and Choi}]{DBLP:conf/emnlp/ChenZC22}
Hung{-}Ting Chen, Michael J.~Q. Zhang, and Eunsol Choi. 2022.
\newblock \href {https://doi.org/10.18653/v1/2022.emnlp-main.146} {Rich knowledge sources bring complex knowledge conflicts: Recalibrating models to reflect conflicting evidence}.
\newblock In \emph{Proceedings of the 2022 Conference on Empirical Methods in Natural Language Processing, {EMNLP} 2022, Abu Dhabi, United Arab Emirates, December 7-11, 2022}, pages 2292--2307. Association for Computational Linguistics.

\bibitem[{Chen et~al.(2023{\natexlab{a}})Chen, Yoon, Ebrahimi, Arik, Pfister, and Jha}]{chen2023adaptation}
Jiefeng Chen, Jinsung Yoon, Sayna Ebrahimi, Sercan~O Arik, Tomas Pfister, and Somesh Jha. 2023{\natexlab{a}}.
\newblock \href {http://arxiv.org/abs/2310.11689} {Adaptation with self-evaluation to improve selective prediction in llms}.

\bibitem[{Chen et~al.(2023{\natexlab{b}})Chen, Gao, Bosselut, Sabharwal, and Richardson}]{chen2023disco}
Zeming Chen, Qiyue Gao, Antoine Bosselut, Ashish Sabharwal, and Kyle Richardson. 2023{\natexlab{b}}.
\newblock \href {http://arxiv.org/abs/2212.10534} {Disco: Distilling counterfactuals with large language models}.

\bibitem[{Deng et~al.(2024)Deng, Zhao, Li, Ng, and Chua}]{deng2024gotcha}
Yang Deng, Yong Zhao, Moxin Li, See-Kiong Ng, and Tat-Seng Chua. 2024.
\newblock \href {http://arxiv.org/abs/2402.15062} {Gotcha! don't trick me with unanswerable questions! self-aligning large language models for responding to unknown questions}.

\bibitem[{Feng et~al.(2024)Feng, Shi, Wang, Ding, Balachandran, and Tsvetkov}]{feng2024dont}
Shangbin Feng, Weijia Shi, Yike Wang, Wenxuan Ding, Vidhisha Balachandran, and Yulia Tsvetkov. 2024.
\newblock \href {http://arxiv.org/abs/2402.00367} {Don't hallucinate, abstain: Identifying llm knowledge gaps via multi-llm collaboration}.

\bibitem[{Guo et~al.(2022)Guo, Schlichtkrull, and Vlachos}]{10.1162/tacl_a_00454}
Zhijiang Guo, Michael Schlichtkrull, and Andreas Vlachos. 2022.
\newblock \href {https://doi.org/10.1162/tacl_a_00454} {{A Survey on Automated Fact-Checking}}.
\newblock \emph{Transactions of the Association for Computational Linguistics}, 10:178--206.

\bibitem[{Hermann et~al.(2015)Hermann, Kočiský, Grefenstette, Espeholt, Kay, Suleyman, and Blunsom}]{hermann2015teaching}
Karl~Moritz Hermann, Tomáš Kočiský, Edward Grefenstette, Lasse Espeholt, Will Kay, Mustafa Suleyman, and Phil Blunsom. 2015.
\newblock \href {http://arxiv.org/abs/1506.03340} {Teaching machines to read and comprehend}.

\bibitem[{Hirschman et~al.(1999)Hirschman, Light, Breck, and Burger}]{hirschman-etal-1999-deep}
Lynette Hirschman, Marc Light, Eric Breck, and John~D. Burger. 1999.
\newblock \href {https://doi.org/10.3115/1034678.1034731} {Deep read: A reading comprehension system}.
\newblock In \emph{Proceedings of the 37th Annual Meeting of the Association for Computational Linguistics}, pages 325--332, College Park, Maryland, USA. Association for Computational Linguistics.

\bibitem[{Ji et~al.(2023)Ji, Lee, Frieske, Yu, Su, Xu, Ishii, Bang, Madotto, and Fung}]{Ji_2023}
Ziwei Ji, Nayeon Lee, Rita Frieske, Tiezheng Yu, Dan Su, Yan Xu, Etsuko Ishii, Ye~Jin Bang, Andrea Madotto, and Pascale Fung. 2023.
\newblock \href {https://doi.org/10.1145/3571730} {Survey of hallucination in natural language generation}.
\newblock \emph{ACM Computing Surveys}, 55(12):1–38.

\bibitem[{Jiang et~al.(2024)Jiang, Sablayrolles, Roux, Mensch, Savary, Bamford, Chaplot, de~las Casas, Hanna, Bressand, Lengyel, Bour, Lample, Lavaud, Saulnier, Lachaux, Stock, Subramanian, Yang, Antoniak, Scao, Gervet, Lavril, Wang, Lacroix, and Sayed}]{jiang2024mixtral}
Albert~Q. Jiang, Alexandre Sablayrolles, Antoine Roux, Arthur Mensch, Blanche Savary, Chris Bamford, Devendra~Singh Chaplot, Diego de~las Casas, Emma~Bou Hanna, Florian Bressand, Gianna Lengyel, Guillaume Bour, Guillaume Lample, Lélio~Renard Lavaud, Lucile Saulnier, Marie-Anne Lachaux, Pierre Stock, Sandeep Subramanian, Sophia Yang, Szymon Antoniak, Teven~Le Scao, Théophile Gervet, Thibaut Lavril, Thomas Wang, Timothée Lacroix, and William~El Sayed. 2024.
\newblock \href {http://arxiv.org/abs/2401.04088} {Mixtral of experts}.

\bibitem[{Jin et~al.(2024)Jin, Cao, Chen, Liu, Jiang, Xu, Li, and Zhao}]{Jin2024TugofWarBK}
Zhuoran Jin, Pengfei Cao, Yubo Chen, Kang Liu, Xiaojian Jiang, Jiexin Xu, Qiuxia Li, and Jun Zhao. 2024.
\newblock \href {https://api.semanticscholar.org/CorpusID:267782658} {Tug-of-war between knowledge: Exploring and resolving knowledge conflicts in retrieval-augmented language models}.

\bibitem[{Kasai et~al.(2024)Kasai, Sakaguchi, Takahashi, Bras, Asai, Yu, Radev, Smith, Choi, and Inui}]{kasai2024realtime}
Jungo Kasai, Keisuke Sakaguchi, Yoichi Takahashi, Ronan~Le Bras, Akari Asai, Xinyan Yu, Dragomir Radev, Noah~A. Smith, Yejin Choi, and Kentaro Inui. 2024.
\newblock \href {http://arxiv.org/abs/2207.13332} {Realtime qa: What's the answer right now?}

\bibitem[{Khashabi(2019)}]{khashabi2019reasoningdriven}
Daniel Khashabi. 2019.
\newblock \href {http://arxiv.org/abs/1908.04926} {Reasoning-driven question-answering for natural language understanding}.

\bibitem[{Khomutova(2014)}]{khomutova}
Tamara Khomutova. 2014.
\newblock \href {https://doi.org/10.1016/j.sbspro.2014.10.174} {Mood and modality in modern english}.
\newblock \emph{Procedia - Social and Behavioral Sciences}, 154.

\bibitem[{Kim et~al.(2023)Kim, Guo, Yu, and Li}]{kim-etal-2023-chatgpt}
Yuheun Kim, Lu~Guo, Bei Yu, and Yingya Li. 2023.
\newblock \href {https://doi.org/10.18653/v1/2023.wassa-1.33} {Can {C}hat{GPT} understand causal language in science claims?}
\newblock In \emph{Proceedings of the 13th Workshop on Computational Approaches to Subjectivity, Sentiment, {\&} Social Media Analysis}, pages 379--389, Toronto, Canada. Association for Computational Linguistics.

\bibitem[{Kojima et~al.(2023)Kojima, Gu, Reid, Matsuo, and Iwasawa}]{kojima2023large}
Takeshi Kojima, Shixiang~Shane Gu, Machel Reid, Yutaka Matsuo, and Yusuke Iwasawa. 2023.
\newblock \href {http://arxiv.org/abs/2205.11916} {Large language models are zero-shot reasoners}.

\bibitem[{Kočiský et~al.(2017)Kočiský, Schwarz, Blunsom, Dyer, Hermann, Melis, and Grefenstette}]{kočiský2017narrativeqa}
Tomáš Kočiský, Jonathan Schwarz, Phil Blunsom, Chris Dyer, Karl~Moritz Hermann, Gábor Melis, and Edward Grefenstette. 2017.
\newblock \href {http://arxiv.org/abs/1712.07040} {The narrativeqa reading comprehension challenge}.

\bibitem[{Kripke(1959)}]{kripke}
Saul~A. Kripke. 1959.
\newblock \href {http://www.jstor.org/stable/2964568} {A completeness theorem in modal logic}.
\newblock \emph{The Journal of Symbolic Logic}, 24(1):1--14.

\bibitem[{Lehnert(1978)}]{Lehnert78-book}
Wendy~G Lehnert. 1978.
\newblock \emph{The Process of Question Answering}.
\newblock Lawrence Erlbaum Associates, Hillsdale, N. J.

\bibitem[{Li et~al.(2022)Li, Rawat, Zaheer, Wang, Lukasik, Veit, Yu, and Kumar}]{li2022large}
Daliang Li, Ankit~Singh Rawat, Manzil Zaheer, Xin Wang, Michal Lukasik, Andreas Veit, Felix Yu, and Sanjiv Kumar. 2022.
\newblock \href {http://arxiv.org/abs/2211.05110} {Large language models with controllable working memory}.

\bibitem[{Liu et~al.(2024)Liu, Wang, Yuan, Chen, and Peng}]{liu2024examining}
Genglin Liu, Xingyao Wang, Lifan Yuan, Yangyi Chen, and Hao Peng. 2024.
\newblock \href {http://arxiv.org/abs/2311.09731} {Examining llms' uncertainty expression towards questions outside parametric knowledge}.

\bibitem[{Longpre et~al.(2021)Longpre, Perisetla, Chen, Ramesh, DuBois, and Singh}]{DBLP:journals/corr/abs-2109-05052}
Shayne Longpre, Kartik Perisetla, Anthony Chen, Nikhil Ramesh, Chris DuBois, and Sameer Singh. 2021.
\newblock \href {http://arxiv.org/abs/2109.05052} {Entity-based knowledge conflicts in question answering}.
\newblock \emph{CoRR}, abs/2109.05052.

\bibitem[{Lu et~al.(2023)Lu, Larcher, and Tran}]{lu2023hybrid}
Guang Lu, Sylvia~B. Larcher, and Tu~Tran. 2023.
\newblock \href {http://arxiv.org/abs/2306.01169} {Hybrid long document summarization using c2f-far and chatgpt: A practical study}.

\bibitem[{Mallen et~al.(2023)Mallen, Asai, Zhong, Das, Khashabi, and Hajishirzi}]{mallen2023trust}
Alex Mallen, Akari Asai, Victor Zhong, Rajarshi Das, Daniel Khashabi, and Hannaneh Hajishirzi. 2023.
\newblock \href {http://arxiv.org/abs/2212.10511} {When not to trust language models: Investigating effectiveness of parametric and non-parametric memories}.

\bibitem[{Matthewson(2016)}]{matthewson_2016}
Lisa Matthewson. 2016.
\newblock \href {https://doi.org/10.1017/CBO9781139236157.019} {\emph{Modality}}, Cambridge Handbooks in Language and Linguistics. Cambridge University Press.

\bibitem[{Mik(2024)}]{mik2024caveat}
Eliza Mik. 2024.
\newblock \href {http://arxiv.org/abs/2403.09163} {Caveat lector: Large language models in legal practice}.

\bibitem[{Neeman et~al.(2022)Neeman, Aharoni, Honovich, Choshen, Szpektor, and Abend}]{neeman2022disentqa}
Ella Neeman, Roee Aharoni, Or~Honovich, Leshem Choshen, Idan Szpektor, and Omri Abend. 2022.
\newblock \href {http://arxiv.org/abs/2211.05655} {Disentqa: Disentangling parametric and contextual knowledge with counterfactual question answering}.

\bibitem[{Pyatkin et~al.(2021)Pyatkin, Sadde, Rubinstein, Portner, and Tsarfaty}]{pyatkin-etal-2021-possible}
Valentina Pyatkin, Shoval Sadde, Aynat Rubinstein, Paul Portner, and Reut Tsarfaty. 2021.
\newblock \href {https://doi.org/10.18653/v1/2021.acl-long.77} {The possible, the plausible, and the desirable: Event-based modality detection for language processing}.
\newblock In \emph{Proceedings of the 59th Annual Meeting of the Association for Computational Linguistics and the 11th International Joint Conference on Natural Language Processing (Volume 1: Long Papers)}, pages 953--965, Online. Association for Computational Linguistics.

\bibitem[{Qian et~al.(2023)Qian, Zhao, and Wu}]{qian2023merge}
Cheng Qian, Xinran Zhao, and Sherry~Tongshuang Wu. 2023.
\newblock \href {http://arxiv.org/abs/2309.08594} {"merge conflicts!" exploring the impacts of external distractors to parametric knowledge graphs}.

\bibitem[{Qin et~al.(2023)Qin, Zhang, Zhang, Chen, Yasunaga, and Yang}]{qin2023chatgpt}
Chengwei Qin, Aston Zhang, Zhuosheng Zhang, Jiaao Chen, Michihiro Yasunaga, and Diyi Yang. 2023.
\newblock \href {http://arxiv.org/abs/2302.06476} {Is chatgpt a general-purpose natural language processing task solver?}

\bibitem[{Rajpurkar et~al.(2018)Rajpurkar, Jia, and Liang}]{rajpurkar2018know}
Pranav Rajpurkar, Robin Jia, and Percy Liang. 2018.
\newblock \href {http://arxiv.org/abs/1806.03822} {Know what you don't know: Unanswerable questions for squad}.

\bibitem[{Shaier et~al.(2024)Shaier, Hunter, and von~der Wense}]{Shaier2024DesiderataFT}
Sagi Shaier, Lawrence~E Hunter, and Katharina von~der Wense. 2024.
\newblock \href {https://api.semanticscholar.org/CorpusID:267334956} {Desiderata for the context use of question answering systems}.
\newblock \emph{ArXiv}, abs/2401.18001.

\bibitem[{Shi et~al.(2023)Shi, Han, Lewis, Tsvetkov, Zettlemoyer, and tau Yih}]{shi2023trusting}
Weijia Shi, Xiaochuang Han, Mike Lewis, Yulia Tsvetkov, Luke Zettlemoyer, and Scott~Wen tau Yih. 2023.
\newblock \href {http://arxiv.org/abs/2305.14739} {Trusting your evidence: Hallucinate less with context-aware decoding}.

\bibitem[{Shuster et~al.(2021)Shuster, Poff, Chen, Kiela, and Weston}]{shuster-etal-2021-retrieval-augmentation}
Kurt Shuster, Spencer Poff, Moya Chen, Douwe Kiela, and Jason Weston. 2021.
\newblock \href {https://doi.org/10.18653/v1/2021.findings-emnlp.320} {Retrieval augmentation reduces hallucination in conversation}.
\newblock In \emph{Findings of the Association for Computational Linguistics: EMNLP 2021}, pages 3784--3803, Punta Cana, Dominican Republic. Association for Computational Linguistics.

\bibitem[{Singhal et~al.(2023)Singhal, Tu, Gottweis, Sayres, Wulczyn, Hou, Clark, Pfohl, Cole-Lewis, Neal, Schaekermann, Wang, Amin, Lachgar, Mansfield, Prakash, Green, Dominowska, y~Arcas, Tomasev, Liu, Wong, Semturs, Mahdavi, Barral, Webster, Corrado, Matias, Azizi, Karthikesalingam, and Natarajan}]{singhal2023expertlevel}
Karan Singhal, Tao Tu, Juraj Gottweis, Rory Sayres, Ellery Wulczyn, Le~Hou, Kevin Clark, Stephen Pfohl, Heather Cole-Lewis, Darlene Neal, Mike Schaekermann, Amy Wang, Mohamed Amin, Sami Lachgar, Philip Mansfield, Sushant Prakash, Bradley Green, Ewa Dominowska, Blaise~Aguera y~Arcas, Nenad Tomasev, Yun Liu, Renee Wong, Christopher Semturs, S.~Sara Mahdavi, Joelle Barral, Dale Webster, Greg~S. Corrado, Yossi Matias, Shekoofeh Azizi, Alan Karthikesalingam, and Vivek Natarajan. 2023.
\newblock \href {http://arxiv.org/abs/2305.09617} {Towards expert-level medical question answering with large language models}.

\bibitem[{Stephany(1986)}]{stephany_1986}
Ursula Stephany. 1986.
\newblock \href {https://doi.org/10.1017/CBO9780511620683.022} {\emph{Modality}}. Cambridge University Press.

\bibitem[{Sugawara et~al.(2021)Sugawara, Stenetorp, and Aizawa}]{sugawara-etal-2021-benchmarking}
Saku Sugawara, Pontus Stenetorp, and Akiko Aizawa. 2021.
\newblock \href {https://doi.org/10.18653/v1/2021.eacl-main.137} {Benchmarking machine reading comprehension: A psychological perspective}.
\newblock In \emph{Proceedings of the 16th Conference of the European Chapter of the Association for Computational Linguistics: Main Volume}, pages 1592--1612, Online. Association for Computational Linguistics.

\bibitem[{Touvron et~al.(2023)Touvron, Martin, Stone, Albert, Almahairi, Babaei, Bashlykov, Batra, Bhargava, Bhosale, Bikel, Blecher, Ferrer, Chen, Cucurull, Esiobu, Fernandes, Fu, Fu, Fuller, Gao, Goswami, Goyal, Hartshorn, Hosseini, Hou, Inan, Kardas, Kerkez, Khabsa, Kloumann, Korenev, Koura, Lachaux, Lavril, Lee, Liskovich, Lu, Mao, Martinet, Mihaylov, Mishra, Molybog, Nie, Poulton, Reizenstein, Rungta, Saladi, Schelten, Silva, Smith, Subramanian, Tan, Tang, Taylor, Williams, Kuan, Xu, Yan, Zarov, Zhang, Fan, Kambadur, Narang, Rodriguez, Stojnic, Edunov, and Scialom}]{touvron2023LLaMA}
Hugo Touvron, Louis Martin, Kevin Stone, Peter Albert, Amjad Almahairi, Yasmine Babaei, Nikolay Bashlykov, Soumya Batra, Prajjwal Bhargava, Shruti Bhosale, Dan Bikel, Lukas Blecher, Cristian~Canton Ferrer, Moya Chen, Guillem Cucurull, David Esiobu, Jude Fernandes, Jeremy Fu, Wenyin Fu, Brian Fuller, Cynthia Gao, Vedanuj Goswami, Naman Goyal, Anthony Hartshorn, Saghar Hosseini, Rui Hou, Hakan Inan, Marcin Kardas, Viktor Kerkez, Madian Khabsa, Isabel Kloumann, Artem Korenev, Punit~Singh Koura, Marie-Anne Lachaux, Thibaut Lavril, Jenya Lee, Diana Liskovich, Yinghai Lu, Yuning Mao, Xavier Martinet, Todor Mihaylov, Pushkar Mishra, Igor Molybog, Yixin Nie, Andrew Poulton, Jeremy Reizenstein, Rashi Rungta, Kalyan Saladi, Alan Schelten, Ruan Silva, Eric~Michael Smith, Ranjan Subramanian, Xiaoqing~Ellen Tan, Binh Tang, Ross Taylor, Adina Williams, Jian~Xiang Kuan, Puxin Xu, Zheng Yan, Iliyan Zarov, Yuchen Zhang, Angela Fan, Melanie Kambadur, Sharan Narang, Aurelien Rodriguez, Robert Stojnic, Sergey Edunov, and Thomas
  Scialom. 2023.
\newblock \href {http://arxiv.org/abs/2307.09288} {Llama 2: Open foundation and fine-tuned chat models}.

\bibitem[{Varshney and Baral(2023)}]{varshney-baral-2023-post}
Neeraj Varshney and Chitta Baral. 2023.
\newblock \href {https://doi.org/10.18653/v1/2023.acl-long.55} {Post-abstention: Towards reliably re-attempting the abstained instances in {QA}}.
\newblock In \emph{Proceedings of the 61st Annual Meeting of the Association for Computational Linguistics (Volume 1: Long Papers)}, pages 967--982, Toronto, Canada. Association for Computational Linguistics.

\bibitem[{Varshney et~al.(2023)Varshney, Parmar, Patel, Handa, Sarkar, Luo, and Baral}]{DBLP:journals/corr/abs-2305-12096}
Neeraj Varshney, Mihir Parmar, Nisarg Patel, Divij Handa, Sayantan Sarkar, Man Luo, and Chitta Baral. 2023.
\newblock \href {https://doi.org/10.48550/arXiv.2305.12096} {Can {NLP} models correctly reason over contexts that break the common assumptions?}
\newblock \emph{CoRR}, abs/2305.12096.

\bibitem[{Wang and Shu(2023)}]{wang2023explainable}
Haoran Wang and Kai Shu. 2023.
\newblock \href {http://arxiv.org/abs/2310.05253} {Explainable claim verification via knowledge-grounded reasoning with large language models}.

\bibitem[{Xie et~al.(2023)Xie, Zhang, Chen, Lou, and Su}]{Xie2023AdaptiveCO}
Jian Xie, Kai Zhang, Jiangjie Chen, Renze Lou, and Yu~Su. 2023.
\newblock \href {https://api.semanticscholar.org/CorpusID:258832560} {Adaptive chameleon or stubborn sloth: Unraveling the behavior of large language models in knowledge clashes}.
\newblock \emph{ArXiv}, abs/2305.13300.

\bibitem[{Zhang et~al.(2023{\natexlab{a}})Zhang, Press, Merrill, Liu, and Smith}]{zhang2023language}
Muru Zhang, Ofir Press, William Merrill, Alisa Liu, and Noah~A. Smith. 2023{\natexlab{a}}.
\newblock \href {http://arxiv.org/abs/2305.13534} {How language model hallucinations can snowball}.

\bibitem[{Zhang et~al.(2023{\natexlab{b}})Zhang, Khalifa, Logeswaran, Lee, Lee, and Wang}]{Zhang2023MergingGA}
Yunxiang Zhang, Muhammad Khalifa, Lajanugen Logeswaran, Moontae Lee, Honglak Lee, and Lu~Wang. 2023{\natexlab{b}}.
\newblock \href {https://api.semanticscholar.org/CorpusID:264426669} {Merging generated and retrieved knowledge for open-domain qa}.
\newblock \emph{ArXiv}, abs/2310.14393.

\bibitem[{Zhou et~al.(2024)Zhou, Hwang, Ren, and Sap}]{Zhou2024RelyingOT}
Kaitlyn Zhou, Jena~D. Hwang, Xiang Ren, and Maarten Sap. 2024.
\newblock \href {https://api.semanticscholar.org/CorpusID:266977353} {Relying on the unreliable: The impact of language models' reluctance to express uncertainty}.
\newblock \emph{ArXiv}, abs/2401.06730.

\bibitem[{Zhou et~al.(2023)Zhou, Zhang, Poon, and Chen}]{zhou2023contextfaithful}
Wenxuan Zhou, Sheng Zhang, Hoifung Poon, and Muhao Chen. 2023.
\newblock \href {http://arxiv.org/abs/2303.11315} {Context-faithful prompting for large language models}.

\end{thebibliography}
\bibliographystyle{acl_natbib}
\clearpage
\appendix

\section*{Appendices}

\section{Basic Prompt Templates} \label{sec:basic}
For ChatGPT and GPT-4:

Prompt 1:
\begin{quote}
 \emph{Text:``\{context\}''}
 
 \emph{Question: ``\{question\}''. Shortest possible answer please.}
 
 \emph{If the question cannot be answered with a single span from the text, return ``None''}
\end{quote}

Prompt 2:
\begin{quote}
 \emph{Text:``\{context\}''}
 
 \emph{Question: ``\{question\}''}
 
 \emph{Important! The answer should be an exact span extracted from the text.}
 \emph{Important! Give the shortest possible answer.}
 \emph{If the question cannot be answered with one span from the text - return ``None'' (and nothing else).}
 \emph{Answer:} 
\end{quote} 

\begin{quote}

For LLaMA 2 and Mixtral:

\textit{Extract from the following context the minimal span word for word that best answers the question. The answer should be the shortest possible. Give the answer in the format as follows:}

                            \textit{Answer: ...}
                            
                            \textit{If the answer is not in the context, the answer should be "None".}
                            
                            \emph{Text:``\{context\}''}
 
                            \emph{Question: ``\{question\}''}
                            
                            \textit{Answer: }
\end{quote}

\section{Instructed Prompting and Chain-of-Thought experiments: Instruction Phrasing}
\label{sec:instr}
Below we cite the instructions which we add to the basic prompts to obtain the so-called instructed prompts.

For gpt-3.5-turbo-0301 with basic prompt 1; for gpt-4-0613 with basic prompts 1 and 2 (instructed prompt and CoT experiments).
\begin{quote}
  \emph{Read the text and answer the question below.}
  
  \emph{Please, ignore your own world knowledge and answer the question based on the text only!}
   
\end{quote}

For gpt-3.5-turbo-0613 with basic prompts 1 and 2 (instructed promptng experiments):
\begin{quote}
  \emph{Review the provided text and address the question that comes afterward.}
  
  \emph{Ensure that your response is solely rooted in the information within the text, disregarding your parametric knowledge.}
   
\end{quote}

For  gpt-3.5-turbo-0301 with basic prompt 2 (instructed prompt experiments); for gpt-3.5-turbo-0613 with basic prompt 2; for LLaMA 2 70B Chat (CoT experiments):
\begin{quote}
  \emph{Read the passage and respond to the question that follows.}
  
  \emph{Please, do not rely on your personal background knowledge; provide your answer solely based on the information presented in the text.}
   
\end{quote}

For LLaMA 2 70B Chat (instruction prompting experiments):
\begin{quote}
  \textit{Examine the text and reply to the question posed below.}
  
  \textit{Your answer should be exclusively informed by the content of the text, without taking into account any prior knowledge or experiences.}
\end{quote}

For Mixtral 8x7B Instruct v0.1 (instructed prompting and CoT experiments); for  gpt-3.5-turbo-0301 and gpt-3.5-turbo-0613 with basic prompt 1 (CoT experiments):
\begin{quote}
  \textit{Review the provided text and address the question that comes afterward.}
  
  \textit{Ensure that your response is solely rooted in the information within the text, disregarding your parametric knowledge.}

\end{quote}

\section{Evaluation Criteria}
\label{sec:eval}
For each prompt/model combination 
we use two setups:
\begin{description}
 \item[Controlled.] We only accept model responses matching a context span or starting with ``None'' (indicating an unanswerable question). Responses with different formats are rejected, and the model is asked to try again. If, after 15 attempts, the model fails to provide a suitable answer, the question is labeled as ``skipped.''
 \item[Uncontrolled.] We do not filter the result format, accepting any response by the model.
\end{description}
The \emph{controlled} setup aligns more closely with the SQuAD extractive QA setting, also typically enhancing evaluation results. The \emph{uncontrolled} setup, deviating from the strict extractive QA paradigm, enables the observation of LLMs' “spontaneous” reactions, which is useful for manual error analysis.

 For evaluation in all the experiments we use the official SQuAD2.0 evaluation script.\footnote{\url{https://github.com/white127/SQUAD-2.0-bidaf/blob/master/evaluate-v2.0.py}.} The evaluation metrics used are exact match and F1. The results presented in the figures are based on the F1 obtained in the controlled setting. The trends for the uncontrolled version are the same.
 
\section{Related Work Table} \label{sec: rel_appendix}
Detailed comparison between this study and other works on knowledge conflict is presented in Table \ref{rel}.
% Please add the following required packages to your document preamble:
% \usepackage[table,xcdraw]{xcolor}
% Beamer presentation requires \usepackage{colortbl} instead of \usepackage[table,xcdraw]{xcolor}

\begin{table*}[t]
\Large
\resizebox{\textwidth}{!}{%
\begin{tabular}{|
>{\columncolor[HTML]{D9EAD3}}l |l|l|l|l|l|l|l|l|l|l|}
\hline
\textbf{work} &
  \cellcolor[HTML]{D9EAD3}\textbf{\begin{tabular}[c]{@{}l@{}}knowledge\\ conflict \\ creation method\end{tabular}} &
  \cellcolor[HTML]{D9EAD3}\textbf{data types} &
  \cellcolor[HTML]{D9EAD3}\textbf{\begin{tabular}[c]{@{}l@{}}context \\ properties\end{tabular}} &
  \cellcolor[HTML]{D9EAD3}\textbf{\begin{tabular}[c]{@{}l@{}}context features \\ considered\end{tabular}} &
  \cellcolor[HTML]{D9EAD3}\textbf{\begin{tabular}[c]{@{}l@{}}semantic\\  modifications\end{tabular}} &
  \cellcolor[HTML]{D9EAD3}\textbf{\begin{tabular}[c]{@{}l@{}}irrelevant \\ contexts\end{tabular}} &
  \cellcolor[HTML]{D9EAD3}\textbf{task types} &
  \cellcolor[HTML]{D9EAD3}\textbf{models} &
  \cellcolor[HTML]{D9EAD3}\textbf{\begin{tabular}[c]{@{}l@{}}number of\\ external\\  contexts\end{tabular}} &
  \cellcolor[HTML]{D9EAD3}\textbf{\begin{tabular}[c]{@{}l@{}}solution/mitigation\\  approach\end{tabular}} \\ \hline
\textbf{Present work} &
  \begin{tabular}[c]{@{}l@{}}half-manual\\  entity substitution;\\ same entity type\end{tabular} &
  \begin{tabular}[c]{@{}l@{}}factual,\\  counterfactual\\  and imaginary\end{tabular} &
  \begin{tabular}[c]{@{}l@{}}simple, \\ straightforward, \\ short contexts\end{tabular} &
   &
  \begin{tabular}[c]{@{}l@{}}focus on\\ semantic\\  modification \\ experiments\end{tabular} &
  \begin{tabular}[c]{@{}l@{}}irrelevant\\  contexts \\ resulting from \\ semantic\\  modifications\end{tabular} &
  extractive QA &
  \begin{tabular}[c]{@{}l@{}}ChatGPT, \\ GPT-4, Llama 2,\\  Mistral\end{tabular} &
  \cellcolor[HTML]{FFFFFF}single &
  \cellcolor[HTML]{FFFFFF}\begin{tabular}[c]{@{}l@{}}using "imaginary"\\  data\\  for testing LLMs'\\  text  understanding \\ abilities\end{tabular} \\ \hline
\textbf{\citet{DBLP:journals/corr/abs-2109-05052}} &
  \begin{tabular}[c]{@{}l@{}}automatic entity \\ substitution;\\ diverse entity types\end{tabular} &
  \begin{tabular}[c]{@{}l@{}}factual vs. \\ counterfactual\end{tabular} &
  \begin{tabular}[c]{@{}l@{}}longer \\  contexts\end{tabular} &
  \begin{tabular}[c]{@{}l@{}}entity\\ popularity,\\  context \\relevance,\\  context \\plausibility\end{tabular} &
  None &
  \begin{tabular}[c]{@{}l@{}}no irrelevant \\ contexts \\ are studied\\  systematically, \\ but some of the\\  retrieved\\  passages \\ can be less \\ relevant \\ than others\end{tabular} &
  \begin{tabular}[c]{@{}l@{}}retrieve-and-read \\ QA \\ and extractive QA\end{tabular} &
  T5, Roberta &
  \cellcolor[HTML]{FFFFFF}single &
  \cellcolor[HTML]{FFFFFF}{\color[HTML]{0D0D0D} \begin{tabular}[c]{@{}l@{}}Training on data \\ augmented\\ with examples \\ modified by \\ entity substitution.\end{tabular}} \\ \hline
\textbf{\citet{neeman2022disentqa}} &
  \begin{tabular}[c]{@{}l@{}}automatic entity\\  substitution; \\ same entity type\end{tabular} &
  \begin{tabular}[c]{@{}l@{}}factual vs. \\ counterfactual\end{tabular} &
  \begin{tabular}[c]{@{}l@{}}longer  \\ contexts\end{tabular} &
   &
  None &
  \begin{tabular}[c]{@{}l@{}}unrelated and \\ empty contexts\end{tabular} &
  \begin{tabular}[c]{@{}l@{}}retrieval QA \\ (using \\ the “gold” \\ passage as\\  the context,\\  assuming an oracle \\ retrieval system.)\end{tabular} &
  T5 &
  \cellcolor[HTML]{FFFFFF}single &
  \cellcolor[HTML]{FFFFFF}\begin{tabular}[c]{@{}l@{}}disentangling \\ knowledge \\ sources\end{tabular} \\ \hline
\textbf{\citet{li2022large}} &
  \begin{tabular}[c]{@{}l@{}}automatic entity \\ substitution; \\ same entity type\end{tabular} &
  \begin{tabular}[c]{@{}l@{}}factual vs. \\ counterfactual\end{tabular} &
  \begin{tabular}[c]{@{}l@{}}longer  \\ contexts\end{tabular} &
   &
  None &
  \begin{tabular}[c]{@{}l@{}}the human labeled \\ "impossible" \\ slice of \\ SQuAD 2.0\end{tabular} &
  \begin{tabular}[c]{@{}l@{}}multiple choice\\  (QASC), \\ Cloze (TReX), \\ extractive \\ (SQuAD) and\\  open domain \\ (TriviaQA) QA\end{tabular} &
  T5, PaLM &
  \cellcolor[HTML]{FFFFFF}single &
  \cellcolor[HTML]{FFFFFF}\begin{tabular}[c]{@{}l@{}}KAFT (knowledge\\  aware finetuning)\\  using data \\ augmented with \\ counterfactual and \\ irrelevant contexts)\end{tabular} \\ \hline
\textbf{\citet{zhou2023contextfaithful}} &
  \begin{tabular}[c]{@{}l@{}}automatic entity\\  substitution, \\ same entity type; \\ relation substitution\\  (for RE)\end{tabular} &
  \begin{tabular}[c]{@{}l@{}}factual vs.\\  counterfactual\end{tabular} &
  \begin{tabular}[c]{@{}l@{}}longer \\  contexts\end{tabular} &
   &
  None &
  unrelated contexts &
  \begin{tabular}[c]{@{}l@{}}reading \\ comprehension,\\  multiple-choice \\ QA, \\ relation extraction\end{tabular} &
  \begin{tabular}[c]{@{}l@{}}Instruct-GPT\\  (text-davinci-003)\end{tabular} &
  \cellcolor[HTML]{FFFFFF}single &
  \cellcolor[HTML]{FFFFFF}\begin{tabular}[c]{@{}l@{}}using prompting\\  strategies\end{tabular} \\ \hline
\textbf{\citet{DBLP:journals/corr/abs-2305-12096}} &
  \cellcolor[HTML]{FFFFFF}{\color[HTML]{0D0D0D} \begin{tabular}[c]{@{}l@{}}manual creation\\  of contexts that\\  follow/break \\ common \\ assumptions\end{tabular}} &
  \begin{tabular}[c]{@{}l@{}}common \\ assumptions vs. \\ broken common \\ assumptions\end{tabular} &
  \begin{tabular}[c]{@{}l@{}}longer\\  contexts;\\  questions\\  that require\\  commonsense\\  reasoning\end{tabular} &
   &
  None &
  \begin{tabular}[c]{@{}l@{}}no irrelevant \\ contexts\end{tabular}&
  \begin{tabular}[c]{@{}l@{}}binary \\ classification\\  (yes-no) questions\end{tabular} &
  \begin{tabular}[c]{@{}l@{}}Flan T5, GPT-3\\  (text-davinci-003), \\ UnifiedQA\end{tabular} &
  \cellcolor[HTML]{FFFFFF}single &
  \cellcolor[HTML]{FFFFFF}n/a \\ \hline
\textbf{\citet{Xie2023AdaptiveCO}} &
  \begin{tabular}[c]{@{}l@{}}replacing entities\\  in the original text +\\  instructing \\ ChatGPT\\  to generate\\  supporting\\ evidence \\ to make the \\ context more \\ coherent\end{tabular} &
  \begin{tabular}[c]{@{}l@{}}parametric \\ memory vs. \\ "counter-\\memory"\end{tabular} &
  \begin{tabular}[c]{@{}l@{}}longer \\ contexts\end{tabular} &
  \begin{tabular}[c]{@{}l@{}}coherence,\\ length,\\  relevance; \\ order \\ of presented \\ evidence; \\ intact vs. \\ fragmented \\ evidence\end{tabular} &
  None &
  \begin{tabular}[c]{@{}l@{}}unrelated contexts \\ (mixed with \\ each other\\  or with relevant \\ ones)\end{tabular} &
  \begin{tabular}[c]{@{}l@{}}multiple-choice \\ QA \\ (options: \\ parametric answer, \\ context answer, \\ "Uncertain")\end{tabular} &
  ChatGPT; GPT-4 &
  \begin{tabular}[c]{@{}l@{}}single and \\ multiple\end{tabular} &
  n/a \\ \hline
\textbf{\citet{DBLP:conf/emnlp/ChenZC22}} &
  \begin{tabular}[c]{@{}l@{}}automatic entity\\  substitution;\\  same entity type\end{tabular} &
  \begin{tabular}[c]{@{}l@{}}parametric \\ answers vs. \\ counter-\\parametric \\ evidence\end{tabular} &
  \begin{tabular}[c]{@{}l@{}}longer\\  contexts\end{tabular} &
   &
  \begin{tabular}[c]{@{}l@{}} negation,\\ changing to \\ future tense,\\ adding modal\\ verb and \\ text infilling\\ (substitution\\  of the sentence \\ root). \\ These \\ perturbations\\ are only applied \\to the original \\ contexts, not to\\  counterfactual \\ ones.\\  Used as an\\ alternative \\ to entity\\ substitution and \\ not on top of it.\end{tabular} &
  \begin{tabular}[c]{@{}l@{}}evidence \\ with different \\ level of relevance \\ mixed together.\end{tabular} &
  extractive QA &
  \cellcolor[HTML]{FFFFFF}FiD, RAG &
  \begin{tabular}[c]{@{}l@{}}multiple \\ conflicting\end{tabular} &
  \cellcolor[HTML]{FFFFFF}\begin{tabular}[c]{@{}l@{}}calibration (creating\\  systems that \\ can detect \\ and abstain from \\ predicting\\ on instances with \\ conflicting evidence)\end{tabular} \\ \hline
\textbf{\citet{mallen2023trust}} &
  None &
  \begin{tabular}[c]{@{}l@{}}parametric \\ knowledge vs.\\  retrieved \\ evidence\end{tabular} &
  \begin{tabular}[c]{@{}l@{}}longer\\  contexts\end{tabular} &
  \begin{tabular}[c]{@{}l@{}}entity\\ popularity;\\  relationship \\ types\end{tabular} &
  None &
  \begin{tabular}[c]{@{}l@{}}no (intentionally \\ used) \\ irrelevant contexts\end{tabular} &
  \begin{tabular}[c]{@{}l@{}}open-domain QA\\  (with and\\  without\\  retrieval-\\ augmentation)\end{tabular} &
  \begin{tabular}[c]{@{}l@{}}GPTNeo, OPT,\\ GPT-3\end{tabular} &
  \cellcolor[HTML]{FFFFFF}\begin{tabular}[c]{@{}l@{}}multiple \\ retrieved \\ contexts\end{tabular} &
  \cellcolor[HTML]{FFFFFF}\begin{tabular}[c]{@{}l@{}}"Adaptive Retrieval" \\ (retrieving \\ non-parametric\\  knowledge only \\ for questions below a \\ “popularity \\ threshold”)\end{tabular} \\ \hline
\textbf{\citet{qian2023merge}} &
  \begin{tabular}[c]{@{}l@{}}automatic entity \\ substitution; \\ same type \\ and type shift;\\  subject or object \\ position\end{tabular} &
  \begin{tabular}[c]{@{}l@{}}factual vs. \\ counterfactual\end{tabular} &
  \begin{tabular}[c]{@{}l@{}}both short\\ (sentence-long)\\  and long \\ (paragraph-long) \\ contexts\end{tabular} &
   \begin{tabular}[c]{@{}l@{}}length, \\ plausibility\end{tabular} &
  None &
  \begin{tabular}[c]{@{}l@{}}irrelevant contexts\\  obtained \\ by both subject \\ and object \\ substitution \\ (with or \\ without type shift)\end{tabular} &
  \begin{tabular}[c]{@{}l@{}}multi-hop QA \\ (with\\  irrelevant or \\ counterfactual\\  distractors of \\ different types \\ introduces \\ at various hops)\end{tabular} &
  \cellcolor[HTML]{FFFFFF}GPT-3.5, MPT-7b &
  \cellcolor[HTML]{FFFFFF}\begin{tabular}[c]{@{}l@{}}single \\ distractors \\ introduced at\\ various hops\end{tabular} &
  \cellcolor[HTML]{FFFFFF}n/a \\ \hline
\textbf{\citet{Zhang2023MergingGA}} &
  \begin{tabular}[c]{@{}l@{}}naturally\\ occurring \\ knowledge \\ conflicts between \\ LLM generated \\ and retrieved \\ passages\end{tabular} &
  \begin{tabular}[c]{@{}l@{}}LLM-generated \\ contexts\\ matched with\\  retrieved \\ contexts by\\  compatibility\end{tabular} &
  \begin{tabular}[c]{@{}l@{}}longer \\ contexts\end{tabular} &
   &
  None &
  \begin{tabular}[c]{@{}l@{}}irrelevant contexts \\ are ranked\\ lower by the \\ compatibillity-\\ based methodology\end{tabular} &
  open-domain QA &
  \cellcolor[HTML]{FFFFFF}\begin{tabular}[c]{@{}l@{}}InstructGPT\\ (text-davinci-002) ; \\ ChatGPT\\ (gpt-3.5-turbo) for \\ LLM-generated\\  passages; \\ FiD for generating\\  final answers\end{tabular} &
  \cellcolor[HTML]{FFFFFF}\begin{tabular}[c]{@{}l@{}}multiple\\  retrieved and \\ LLM-generated \\ contexts ranked\\ by mutual\\ compatibility\end{tabular} &
  \cellcolor[HTML]{FFFFFF}\begin{tabular}[c]{@{}l@{}}COMBO:\\ combining \\retrieved and\\  generated knowledge\\  into compatible pairs, \\ which are then fed into \\ a FiD-based reader, \\ which processes \\ passage pairs, sorted\\  by compatibility\\ scores, to \\generate\\  the final answer.\end{tabular} \\ \hline
\textbf{\citet{Jin2024TugofWarBK}} &
  \cellcolor[HTML]{FFFFFF}\begin{tabular}[c]{@{}l@{}}Distilling \\ counterfactuals\\  with LLMs \\ \citep{chen2023disco}\end{tabular} &
  \begin{tabular}[c]{@{}l@{}}LLM-generated\\  truthful,\\  misleading and\\  irrelevant\\  contexts\end{tabular} &
  \begin{tabular}[c]{@{}l@{}}longer \\ contexts\end{tabular} &
  \begin{tabular}[c]{@{}l@{}}availability of \\ knowledge, \\ frequency of \\ evidence, \\ consistency with\\  the LLM's\\  internal memory,\\  number of \\ conflicting\\  hops (in\\ multi-hop \\ scenarios)\end{tabular} &
  None &
  \begin{tabular}[c]{@{}l@{}}LLM-generated \\ irrelevant\\ contexts\end{tabular} &
  open-book QA &
  \cellcolor[HTML]{FFFFFF}\begin{tabular}[c]{@{}l@{}}FLAN-T5-XL 3B, \\ FLAN-\\ T5-XXL 11B,\\  FLAN-UL2 20B, \\ Baichuan2 7B, \\ Baichuan2 13B, \\ LLaMA2 7B, \\ LLaMA2 13B.\end{tabular} &
  \cellcolor[HTML]{FFFFFF}\begin{tabular}[c]{@{}l@{}}multiple \\ contexts\end{tabular} &
  \cellcolor[HTML]{FFFFFF}{\color[HTML]{1F1F1F} \begin{tabular}[c]{@{}l@{}}Conflict-Disentangle \\ Contrastive \\ Decoding (CD2)\end{tabular}} \\ \hline
\end{tabular}

}

\caption{Comparison between this study and other works on knowledge conflict}
\label{rel}
\end{table*}

\end{document}